\documentclass[10pt,twocolumn,letterpaper]{article}

\usepackage{cvpr}
\usepackage{times}
\usepackage{epsfig}
\usepackage{graphicx}
\usepackage{amsmath}
\usepackage{amssymb}

\usepackage{relsize}
\usepackage{color}

\newcommand{\boldhead}[1]{\vspace{0.05in}\noindent\textbf{#1.}}

\newcommand{\comment}[1]{}

\newcommand{\supi}[1]{{#1}^{ {\left( i \right)} }}
\newcommand{\round}[1]{ \operatorname{round}\left( #1 \right) }

\renewcommand{\baselinestretch}{.95}
\usepackage[pagebackref=true,breaklinks=true,letterpaper=true,colorlinks,bookmarks=false]{hyperref}

\cvprfinalcopy % *** Uncomment this line for the final submission

\def\cvprPaperID{504} % *** Enter the CVPR Paper ID here

\begin{document}

%%%%%%%%% TITLE
\title{Boundary Cues for 3D Object Shape Recovery}

\author{Kevin Karsch${}^1$ \hspace{6mm}
Zicheng Liao${}^1$ \hspace{6mm}
Jason Rock${}^1$ \hspace{6mm}
Jonathan T. Barron${}^2$ \hspace{6mm}
Derek Hoiem${}^1$ \hspace{6mm}
\\
\centerline{${}^1$University of Illinois at Urbana-Champaign\hspace{10mm} ${}^2$University of California, Berkeley \hspace{0mm}}
\\
\centerline{\tt \small \{karsch1, liao17, jjrock2, dhoiem\}@illinois.edu \hspace{10mm} barron@eecs.berkeley.edu \hspace{12mm}}
}

\maketitle
% \thispagestyle{empty}

%%-----------------------------------------------
%\begin{figure}[t]
%\vspace{-2in}
%\begin{minipage}{\textwidth}
%Input image/labels\\
%\centerline{ \includegraphics[width=\linewidth]{fig/teaser.png}}
%\caption{We }
%\label{fig:sample_shapes}
%\end{minipage}
%\end{figure}
%%-----------------------------------------------

%%%%%%%%%%%%%%%%%%%%%%%%%%%%%%%%%%%%%%%%%%%%%%%%
\begin{abstract}
Early work in computer vision considered a host of geometric cues for both shape reconstruction~\cite{MalikThesis} and recognition~\cite{Roberts}. However, since then, the vision community has focused heavily on shading cues for reconstruction~\cite{barron2012eccv}, and moved towards data-driven approaches for recognition~\cite{felzenszwalb2010pami}. In this paper, we reconsider these perhaps overlooked ``boundary'' cues (such as self occlusions and folds in a surface), as well as many other established constraints for shape reconstruction. In a variety of user studies and quantitative tasks, we evaluate how well these cues inform shape reconstruction (relative to each other) in terms of both shape quality and shape recognition. Our findings suggest many new directions for future research in shape reconstruction, such as automatic boundary cue detection and relaxing assumptions in shape from shading (e.g. orthographic projection, Lambertian surfaces).
\end{abstract}

%%%%%%%%%%%%%%%%%%%%%%%%%%%%%%%%%%%%%%%%%%%%%%%%
\thispagestyle{empty}
\section{Introduction}
3D object shape is a major cue to object category and function.  Early approaches to object recognition~\cite{Roberts} considered shape reconstruction as the first step. As data-driven approaches to recognition became popular, researchers began to represent shape implicitly through weighted image gradient features, rather than explicitly through reconstruction~\cite{Mundy06}.  The best current approaches recognize objects with mixtures of gradient-based templates.  Were the early researchers misguided to focus on explicit shape representation?

We have good reason to reconsider the importance of 3D shape.  A study by Hoiem et al.~\cite{hoiem2012eccv} provides some evidence that gradient-based features are a limiting factor in object detection performance.  Distinct architectures~\cite{felzenszwalb2010pami, vedaldi09iccv} whose main commonality is gradient-based feature representations have very similar performance characteristics.  The study also suggests that performance may be limited by heavy-tailed appearance distributions of object categories.  For example, projected dog shapes may vary due to pose, viewpoint, and high intraclass variation.  Because many examples are required to learn which boundaries are reliable (i.e., correspond to shape), dogs of unusual variety, pose, or viewpoint are poorly classified.  Representations based on 3D shape would enable more sample-efficient category learning through viewpoint robustness and a reduced need to learn stable boundaries through statistics. Beyond interest in object categorization, ability to recover 3D shape is important for inferring object pose and affordance and for manipulation tasks.

The importance of shape is clear, but there are many mysteries to be solved before we can recover shape.  What cues are important?  What errors in 3D shape are important?  How do we recover shape cues from an image?  How do we encode and use 3D shape for recognition?  In this paper, we focus on improving our understanding of the importance of boundary shape cues for 3D shape reconstruction and recognition.  In particular, we consider boundaries due to object silhouette, self-occlusion (depth discontinuity) and folds (surface normal discontinuity).  We also consider
%figure/ground cues for occlusion boundaries and
cues for whether boundaries are soft (extrema of curved surface) or sharp.  We evaluate on a standard 3D shape dataset and a selection of PASCAL VOC object images.  On the standard dataset, reconstructions using various cues are compared via metrics of surface normal and depth accuracy.  On the VOC dataset, we evaluate reconstructions qualitatively and in terms of how well people and computers can categorize objects given the reconstructed shape.

%-----------------------------------------------
\begin{figure*}
\begin{minipage}{\textwidth}
\begin{minipage}{.14\linewidth}\centerline{Input image/labels}\end{minipage}
\begin{minipage}{.139\linewidth}\centerline{silh}\end{minipage}
\begin{minipage}{.139\linewidth}\centerline{+selfocc}\end{minipage}
\begin{minipage}{.139\linewidth}\centerline{+folds}\end{minipage}
\begin{minipage}{.139\linewidth}\centerline{+occ+folds}\end{minipage}
\begin{minipage}{.135\linewidth}\centerline{+shading}\end{minipage}
\begin{minipage}{.135\linewidth}\centerline{+shading+occ+folds}\end{minipage}\\
\centerline{ \includegraphics[width=\linewidth]{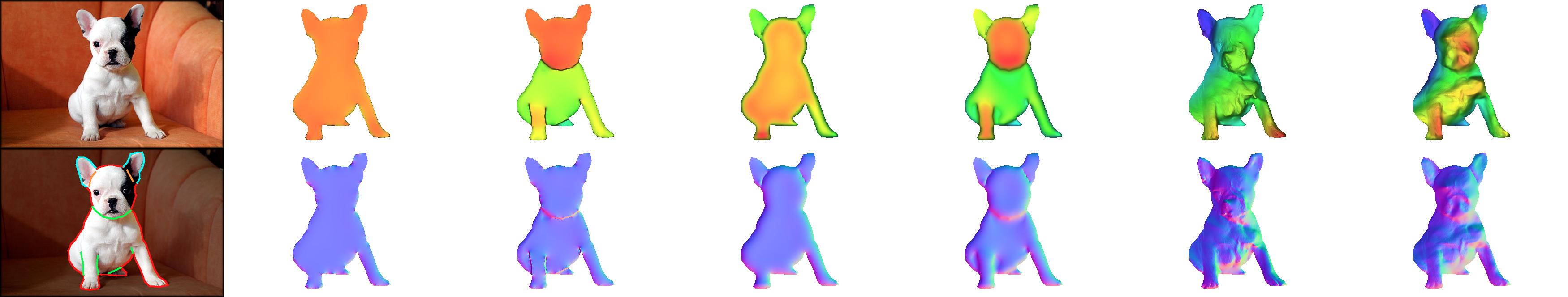}}
\caption{For a given input image, we hand-label geometric cues including: smooth silhouette contour (red), sharp silhouette contour (cyan), self occlusions (green), and folds (orange). We then use various combinations of these cues (as well as appearance-based cues) to obtain different shape reconstructions (see Sec~\ref{sec:eval}). We evaluate these reconstructions in a variety of tasks in order to find which set(s) of cues may be most beneficial for reconstructing shapes.}
\label{fig:sample_shapes}
\end{minipage}
\end{figure*}
%-----------------------------------------------

\boldhead{Contributions}
Our main contribution is to evaluate the importance of various boundary and shading cues for shape reconstruction and shape-based recognition. We extend Barron and Malik's shape from shading and silhouette method~\cite{barron2012eccv} to include interior occlusions with figure/ground labels, folds, and sharp/soft boundary labels.  The standard evaluation is based on depth error, surface normal, shading, or reflectance on the MIT Intrinsic Image dataset.  We also introduce perceptual and recognition-based measures of reconstruction quality for the PASCAL VOC dataset (Fig~\ref{fig:sample_shapes} shows one example of the types of reconstructions we evaluate, and the annotation required by our algorithm). These experiments are important because they tests reconstruction of typical objects, such as cats and boats, with complex shapes and materials in natural environments, and because it can provide insight into which errors matter. Furthermore, much work has gone into shape-based representations for recognition, focusing on the cues provided by the silhouette (e.g. Ferrari et al.~\cite{ferrari:hal-00204002}).  Our findings suggest a 3D representation that incorporates interior occlusions and folds might benefit such existing systems.

\boldhead{Limitations}  Our study is a good step towards understanding shape reconstruction in the context of recognition, but we must leave several aspects of this complex problem unexplored.  First, we assume boundary cues are provided.  Eventually, we will want automatic recovery of shape cues and reconstruction algorithms that handle uncertainty.  Second, cues such as ground contact points and object-level shape priors are useful but not investigated.  Third, we assume an orthographic projection which can be a poor assumption for large objects, such as busses or trains.  Finally, we recover depth maps, which provides a 2.5D reconstruction, rather than a full 3D reconstruction.

%%%%%%%%%%%%%%%%%%%%%%%%%%%%%%%%%%%%%%%%%%%%%%%%

\section{Cues for object reconstruction}
We focus on reconstructing shape from geometric cues, revisiting early work on reconstructing shape from line drawings~\cite{MalikThesis,MalikMaydan}. Through human labeling, we collect information about an object's {\it silhouette}, {\it self-occlusions}, and {\it folds} in the surface. Since appearance can be a helpful factor in determining shape, we also investigate the benefit of shading cues using the shape-from-shading priors of Barron and Malik~\cite{barron2012eccv}. Figure~\ref{fig:sample_shapes} shows reconstructions using each of these cues.

To reconstruct shapes, we extend the continuous optimization framework of Barron and Malik by building in additional constraints on the surface. Following Barron and Malik's notation, we write $Z$ for the surface (represented by a height field viewed orthographically), and $N : \mathbb{R} \rightarrow \mathbb{R}^3 $ as the function that takes a height field to surface normals (component-wise; $N = (N^x, N^y, N^z)$). We use a coordinate system such that $x$ and $y$ vary in the image plane, and negative $z$ is in the viewing direction.

Extending Barron and Malik's continuous optimization framework, we write our optimization problem as:
\begin{eqnarray}
\label{eq:objective}
\displaystyle \underset{ Z, R, L }{\operatorname{minimize}} \!\!&\!&\!\! \delta_\mathit{sfc} f_\mathit{sfc}( Z ) + \delta_\mathit{selfocc} f_\mathit{selfocc}(Z) \nonumber \\
\!\!&\!&\!\!  + \delta_\mathit{fold} f_\mathit{fold}(Z) + \delta_\mathit{reg} f_\mathit{reg}(Z) \nonumber \\
\!\!&\!&\!\!   + \delta_\mathit{sfs} (g(R)  +  h(L)) \nonumber \\
\displaystyle \operatorname{subject\;to} \!\!&\!&\!\!  c_\mathit{sfs}(Z,R,L) = 0,
\end{eqnarray}
where $f_{{}^*}$ and $c_{\mathit{sfs}}$ are sub-objective and constraint functions, $g(R)$ and $h(L)$ are priors on reflectance and illumination, and $\delta_{{}^*}$ are the weights that determine their influence. In the remainder of the section, we describe each of these functions/constraints.

\boldhead{Silhouette} The silhouette is rich with shape information, both perceptually and geometrically~\cite{koenderink1984does}.  At the occluding contour of an object, the surface is tangent to all rays from the vantage point, unless however there is a discontinuity in surface normals across the visible and non-visible regions of the object (e.g. the edges of a cube).  We treat these two cases separately, labeling parts of the silhouette as {\it smooth} if the surface normal should lie perpendicular to both the viewing direction and image silhouette, and {\it sharp} otherwise\footnote{It is also common notation to denote smooth boundaries as ``limbs'' and sharp boundaries as ``edges'' or ``cuts''}. In the case of a smooth silhouette contour, the $z$-component of the normal is $0$, and the $x$ and $y$ components are normal to the silhouette (i.e. perpendicular to the silhouette's tangent in 2D). Denoting $(n^x,n^y)$ as normals of the silhouette contour, and $C_{\mathit{smooth}}$ as the set of pixels labelled as the smooth part of the silhouette,
we write the silhouette constraint as:
\begin{equation}
f_{\mathit{sfc}}(Z) = \sum_{i\in C_{\mathit{smooth}}} \sqrt{ (N_i^x(Z)-n_i^x)^2 +  (N_i^y(Z)-n_i^y)^2 }.
\label{eq:sfc}
\end{equation}
This is the most typical constraint used in shape-from-contour algorithms (hence the notation $f_{\mathit{sfc}}$), and is identical to that used by Barron and Malik, with the notable exception that we only enforce the constraint when the silhouette is not sharp. If the silhouette is labelled sharp, there is no added constraint.

\boldhead{Self-occlusions} Self-occlusions can be thought of in much the same way as the silhouette. The boundary of a self-occlusion implies a {\it discontinuity in depth}, and thus the surface along the foreground boundary should be constrained to be tangent to the viewing direction. Besides knowing a self occlusion boundary, it is also mandatory to know which side of the contour is in front of the other (figure and ground labels). With this information, we impose additional surface normal constraints along self occlusion boundaries ($C_{\mathit{selfocc}}$):
\begin{equation}
f_{\mathit{selfocc}}(Z) = \hspace{-2mm} \sum_{i\in C_{\mathit{selfocc}}} \sqrt{ (N_i^x(Z)-n_i^x)^2 +  (N_i^y(Z)-n_i^y)^2 }.
\label{eq:selfocc}
\end{equation}
Notice that there is no explicit constraint to force the height of the foreground to be greater than that of the background; however, by constraining the foreground normals to be pointing outward and perpendicular to the viewing direction, the correct effect is achieved. This is due in part because we enforce integrability of the surface (since height is directly optimized).

\boldhead{Folds} A fold in the surface denotes a {\it discontinuity in surface normals} across a contour along the object, e.g. edges where faces of a cube meet. Folds can be at any angle (e.g. folds on a cube are at $90^\circ$, but this is not always the case), and can be convex (surface normals pointing away from each other) or concave (surface normals pointing towards each other). Our labels consist of fold contours and also a flag denoting whether the given fold is convex or concave. We did not annotate exact fold orientation as this task is susceptible to human error and tedious.

We incorporate fold labels by adding another term to our objective function, developed using intuition from Malik and Maydan~\cite{MalikMaydan}. The idea is to constrain normals at pixels that lie across a fold to have convex or concave orientation (depending on the label), and to be oriented consistently in the direction of the fold. Let $\mathbf{u}=(\mathbf{u}_x,\mathbf{u}_y,0)$ be a fold's tangent vector in the image plane, and $N_i^\ell, N_i^r$ as two corresponding normals across pixel $i$ in the fold contour $C$. We write the constraint as
\begin{eqnarray}
f_{\mathit{fold}}(Z) = \sum_{i \in C} \max(0, \epsilon - (N_i^\ell \times N_i^r) \cdot \mathbf{u}),
\end{eqnarray}
and set $\epsilon = {1 \over \sqrt{2}}$ (additional details can be found in the appendix).

\boldhead{Regularization priors} Because we only have constraints at a sparse set of points on the surface, we incorporate additional terms to guide the optimization to a plausible result. Following Barron and Malik, we impose one prior that prefers the flattest shape within the bas-relief family ($f_f$), and another that minimizes change in mean curvature ($f_k$):

\begin{align}
f_f(Z) &= -\sum_{i \in \mathit{pixels}} \log\left( N^z_{i}(Z) \right), \\
f_k(Z) &= \sum_{i \in \mathit{pixels}} \sum_{j \in \mathit{neighbors}(i)} c\left(  H(Z)_i - H(Z)_j \right),\\
f_{\mathit{reg}}(Z) &= \lambda_f f_f(Z) + \lambda_k f_k(Z),
\end{align}
where $c(\cdot)$ is the negative log-likelihood of a Gaussian scale mixture, $H(\cdot)$ computes mean curvature, and the neighbors are in a 5x5 window around $i$. For all of our reconstructions, we set $\lambda_f = \lambda_k = 1$ (see~\cite{barron2012eccv} for implementation details).

\boldhead{Shading} We use the albedo and illumination priors of Barron and Malik to incorporate shading cues into our reconstructions. Summarizing these priors, we encourage albedo to be be piecewise smooth over space. Illumination is parameterized by second order spherical harmonics (9 coefficients per color channel), and is encouraged to match a Gaussian fit to real world spherical harmonics (regressed from an image based lighting dataset\footnote{\url{http://www.hdrlabs.com/sibl}}). For brevity, we denote priors on reflectance as $g(R)$, and priors on illumination as $h(L)$, where $R$ is log-diffuse reflectance (log-albedo) and $L$ is the 27-dimensional RGB spherical harmonic coefficient vector. We refer the reader to~\cite{barron2012eccv,barron2012cvpr} for further details.

Jointly estimating shape along with albedo and illumination requires an additional constraint that forces a rendering of the surface to match the input image. Assuming Lambertian reflectance and disregarding occlusions, our rendering function is simply reflectance multiplied by shading (or in log space, log-reflectance plus log-shading). Denoting $I$ as the log-input image, $R$ as log-diffuse reflectance (log-albedo), and $S(Z,L)$ as the log-shaded surface $Z$ under light $L$, we write the shape-from-shading constraint as:
\begin{equation}
c_{\mathit{sfs}}(Z,R,L) = R + S(Z,L) - I.
\end{equation}
We emphasize that $R$, $I$, and $S(\cdot)$ are all in log-space, as is done in~\cite{barron2012eccv} which allows us to write the rendering constraint in additive fashion.

%-----------------------------------------------
\begin{figure*}
\begin{center}
\begin{minipage}{.98\linewidth}
\begin{minipage}{.13\linewidth}\centerline{Input +}\centerline{annotations}\end{minipage}
\hspace{5mm}
\begin{minipage}{.26\linewidth}\centerline{\bf +occ+folds} \centerline{\small View 1 \hspace{10mm} View 2} \end{minipage}
\hspace{2mm}
\begin{minipage}{.26\linewidth}\centerline{\bf +shading} \centerline{\small View 1 \hspace{10mm} View 2} \end{minipage}
\hspace{2mm}
\begin{minipage}{.26\linewidth}\centerline{\bf +shading+occ+folds} \centerline{\small View 1 \hspace{10mm} View 2} \end{minipage}
\vspace{1mm}\\
\includegraphics[width=\linewidth]{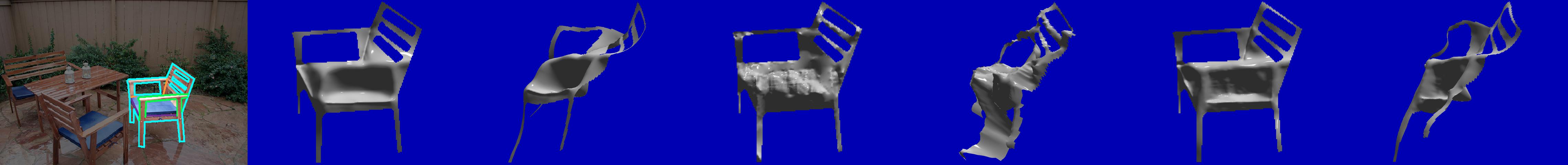}\\
\includegraphics[width=\linewidth]{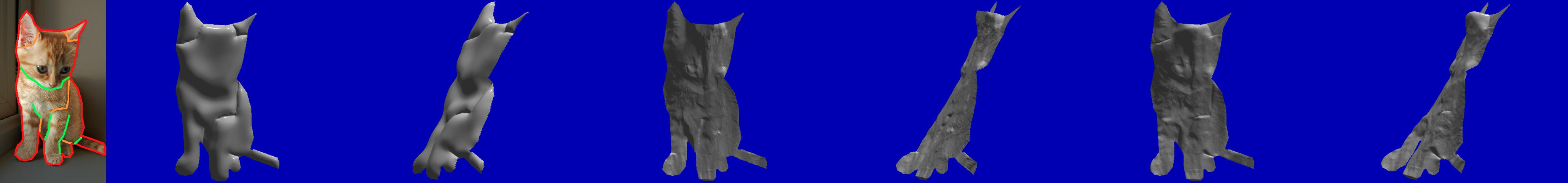}\\
\includegraphics[width=\linewidth]{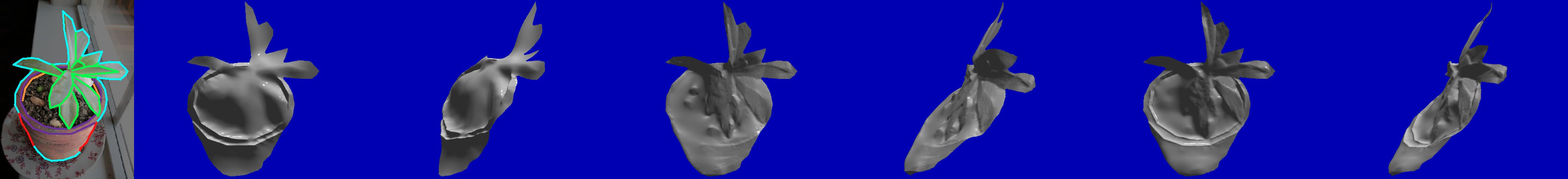}\\
\includegraphics[width=\linewidth]{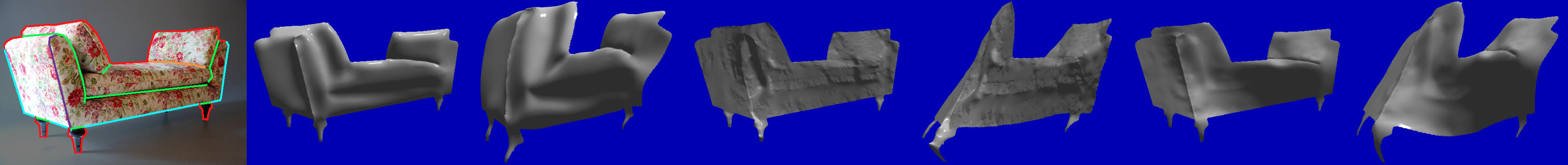}\\
\includegraphics[width=\linewidth]{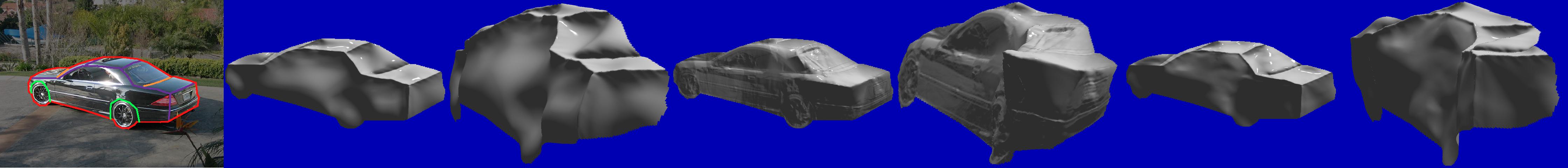}\\
\includegraphics[width=\linewidth]{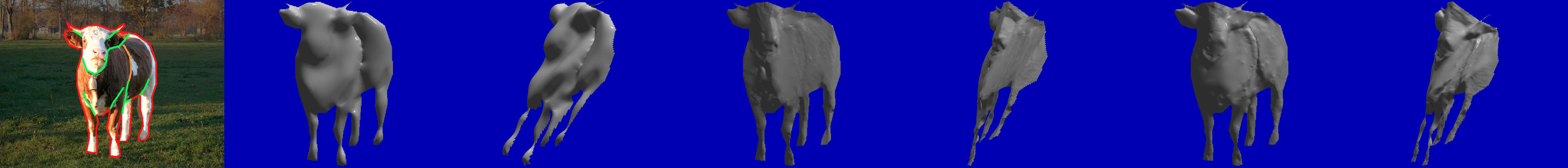}
\end{minipage}
\vspace{-0mm}
\end{center}
\caption{Several annotations and shape reconstructions used in our analyses. The annotated images (left) include: smooth silhouette contour (red), sharp silhouette contour (cyan), self occlusions (green), and folds (orange). In each row, we show the input image (with geometric labels), and the results of three reconstruction algorithms. For each algorithm, two views of the shape are shown (frontal on left, heavily rotated view on right). Notice that the reconstructed shapes look generally good frontally.  Rotated views expose that shape estimates often err towards being too flat (especially with the cow or potted plant).  This paper is the first that we know of to provide a rigorous analysis of shape reconstruction on typical objects in consumer photographs (e.g. outside of a lab setting).}
\label{fig:results}
\end{figure*}
%-----------------------------------------------

%%%%%%
\subsection{Optimization}
To estimate a shape given a set of labels, we solve the optimization problem in Eq.~\ref{eq:objective} using the multiscale optimization technique introduced by Barron and Malik~\cite{barron2012eccv}. Notice that shading cues are only incorporated if $\delta_{\mathit{sfs}}>0$; otherwise, our reconstructions rely purely on geometric information.

\boldhead{Setting the weights ($\delta$)} Throughout our experiments, we choose each weight to be binary for two reasons. For one, each term in the objective should have an equal weighting for a fair comparison, otherwise one cue may dominate others. Second, learning these weights requires a dataset of ground truth shapes, and we have good reason to believe that weights learned from existing datasets (e.g. the MIT Intrinsic Image dataset~\cite{grosse09intrinsic}) will not generalize to shapes found in the VOC dataset (e.g. more geometric detail on VOC shapes). Furthermore, we ran the MIT-learned parameters on several of the VOC images, and noticed only slight perceptual differences in results.

%-----------------------------------------------
\begin{figure}[t]
\centering
\includegraphics[width=0.95\linewidth]{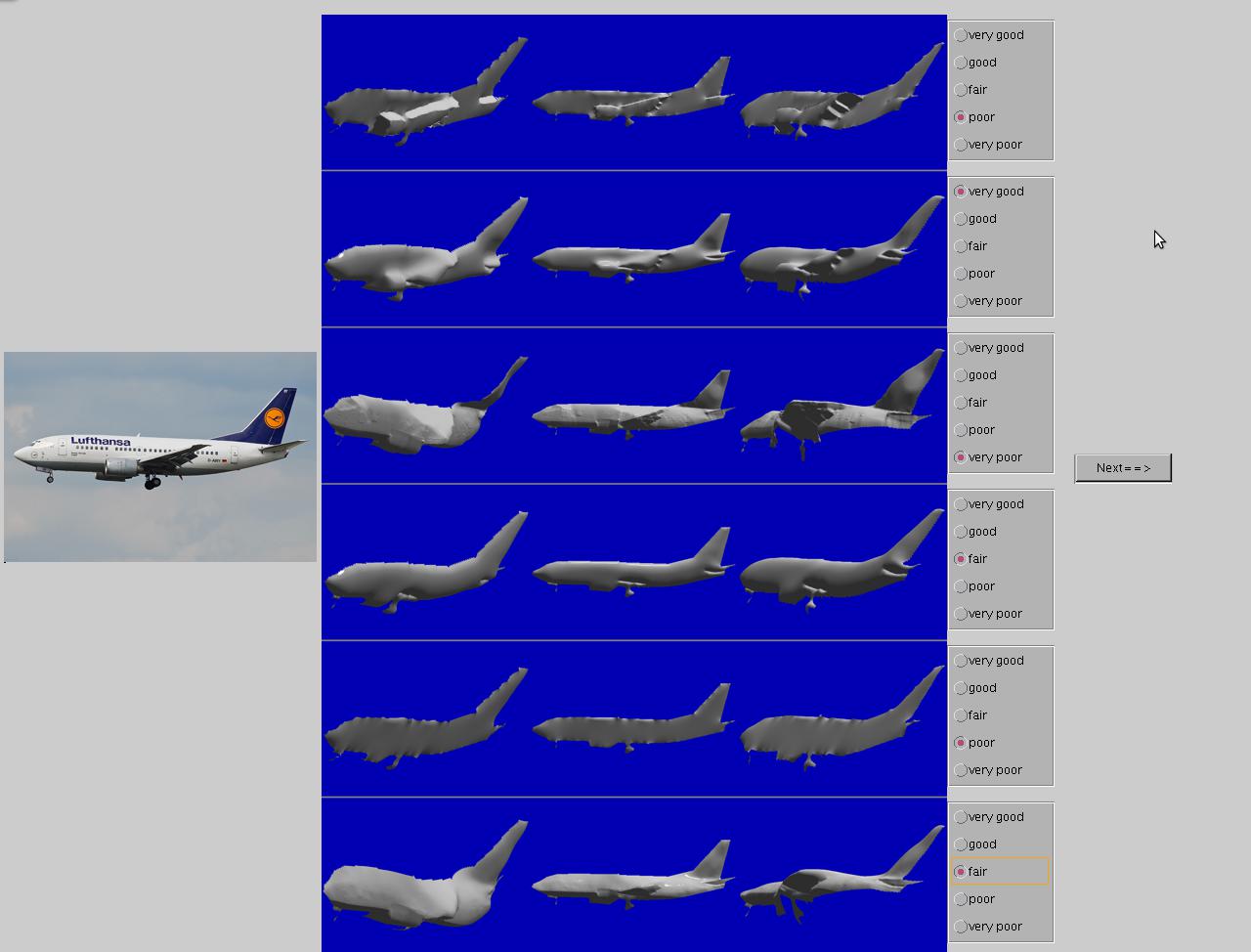}
\caption{User study interface for qualitative rating. Left: source image of the object. Middle: shape visualizations. Each row is the result of one algorithm (silh, +selfocc, etc) in random order visualized in three view angles: upper left, frontal, bottom right (from left to right). The participant rates each row as a whole.}
\label{fig:interface1}
\end{figure}
%-----------------------------------------------

%%%%%%%%%%%%%%%%%%%%%%%%%%%%%%%%%%%%%%%%%%%%%%%%
\section{Evaluation of shape and appearance cues}
\label{sec:eval}

In this section, we examine each of the cues used in our shape reconstruction method, and hope to find a cue or set of cues that lead to better shape estimates (qualitatively, and in terms of recognition ability).  Our objective function (Eq~\ref{eq:objective}) allows us to easily produce shape reconstructions for various combinations of cues by turning ``on'' and ``off'' different cues; equivalently, setting the corresponding weights to 1 (on) or 0 (off). We use six different cue combinations to see which cue or set of cues contribute most to a better reconstruction. These six combinations are:
\begin{itemize}
\item {\bf silh}: Priors on silhouette shape and surface smoothness; i.e. shape-from-contour constraints ($\delta_{\mathit{sfc}} = 1$).
\item {\bf +selfocc}: Silhouette and self occlusion constraints ($\delta_{\mathit{sfc}} =  \delta_{\mathit{selfocc}} = 1$).
\item {\bf +folds}: Silhouette and fold constraints ($\delta_{\mathit{sfc}} =  \delta_{\mathit{folds}} = 1$).
\item {\bf +occ+folds}: Silhouette,  self occlusion and fold constraints ($\delta_{\mathit{sfc}} =  \delta_{\mathit{selfocc}} = \delta_{\mathit{folds}} = 1$).
\item {\bf +shading}: Shape-from-shading as in~\cite{barron2012eccv}; includes silh ($\delta_{\mathit{sfc}} =  \delta_{\mathit{sfs}} = 1$).
\item {\bf +shading+occ+folds}: SFS with self occlusion and fold constraints ($\delta_{\mathit{sfc}} =  \delta_{\mathit{sfs}} = \delta_{\mathit{selfocc}} = \delta_{\mathit{folds}} = 1$).
\end{itemize}

We will refer to these as separate {\it algorithms} for the remainder of the paper, and Fig~\ref{fig:sample_shapes} shows an example reconstruction for each of these algorithms.  Note that silh cues are present in each algorithm (hence the `+' prefix).
%\todo{We will refer to these as separate {\it algorithms} for the remainder of the paper. Fig~\ref{fig:sample_shapes} shows an example reconstruction for each of these algorithms, and Table~\ref{tab:combinations} shows the weights ($\delta$) used by each algorithm.}

To find which cues are most critical for recovering shape, we evaluate each algorithm on a variety of tasks that measure {\it shape quality} and {\it shape recognition}.  We first evaluate the performance of the six algorithms on the VOC 2012 dataset. We selected 17 of the VOC categories out of the 20 (we exclude ``bicycle'', ``motorbike'' and ``person'' since we found these objects difficult to label by hand). Each class has 10 examples. Since we do not have ground truth shape for VOC objects, we conduct two user studies to evaluate qualitative performance: \emph{qualitative rating} and \emph{shape-based recognition}. Next, we evaluated the different algorithms using existing automatic recognition techniques, and compare them to the results of using RGB features (alone) and RGB+depth features. Finally, we ran a quantitative comparison of depth and surface normals using the MIT depth dataset. The remainder of this section details our results for each of these tasks, split under headings concerning shape quality and shape recognition.

%-----------------------------------------------
\begin{figure}[t]
\centering
\includegraphics[width=0.95\linewidth]{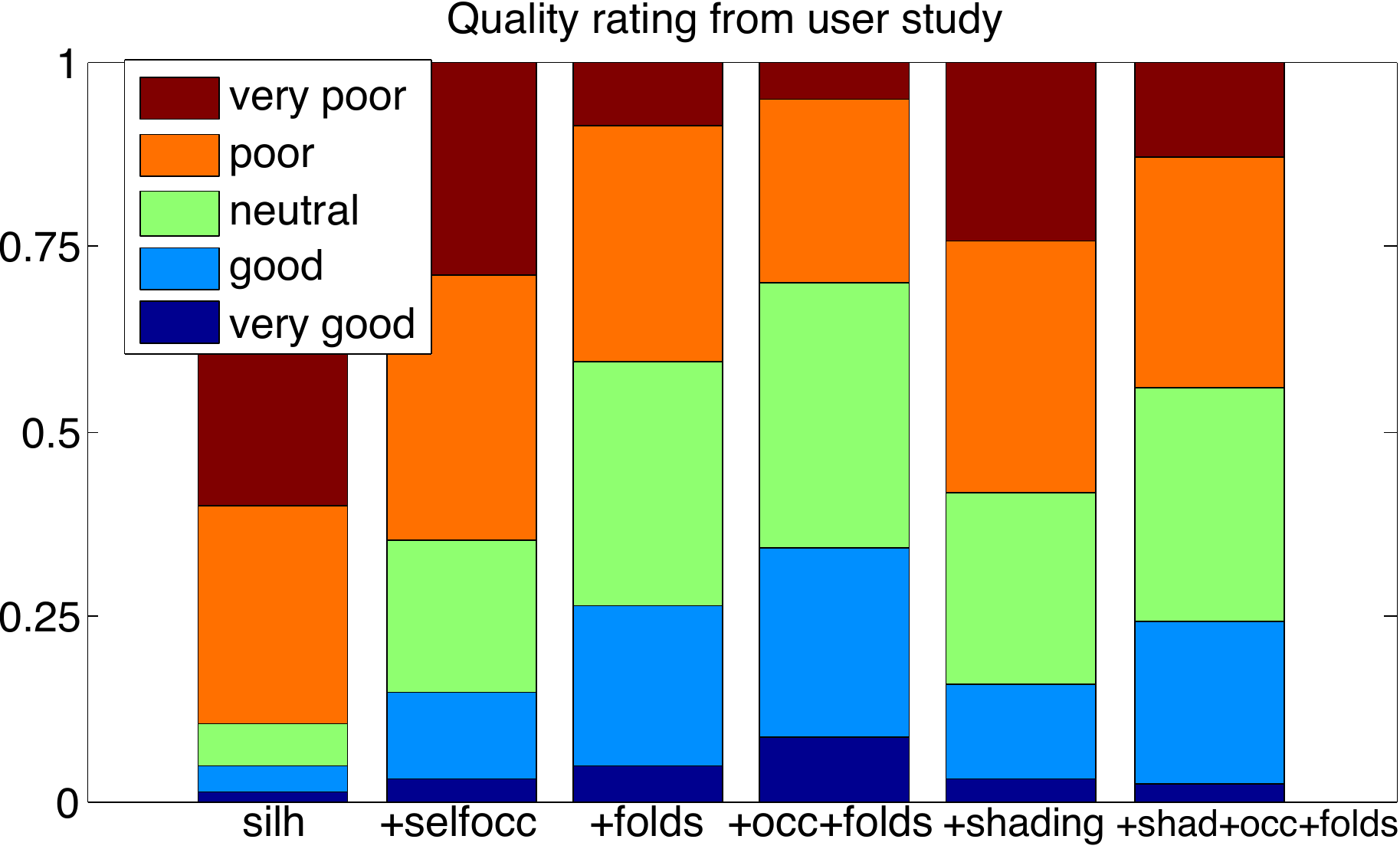}
\caption{For each algorithm, we show the percentage of times a certain rating was assigned to it during the qualitative rating user study. +occ+folds had the highest average rating, followed closely by +folds and +shading+occ+folds. Notice however that there is still much room for improvement, since the best-rated method (+occ+folds) was only chosen as ``good'' or ``very good'' less than 30\% of the time.}
\label{fig:summary}
\end{figure}
%-----------------------------------------------

%-----------------------------------------------
\begin{figure}
\centering
\includegraphics[width=1\linewidth]{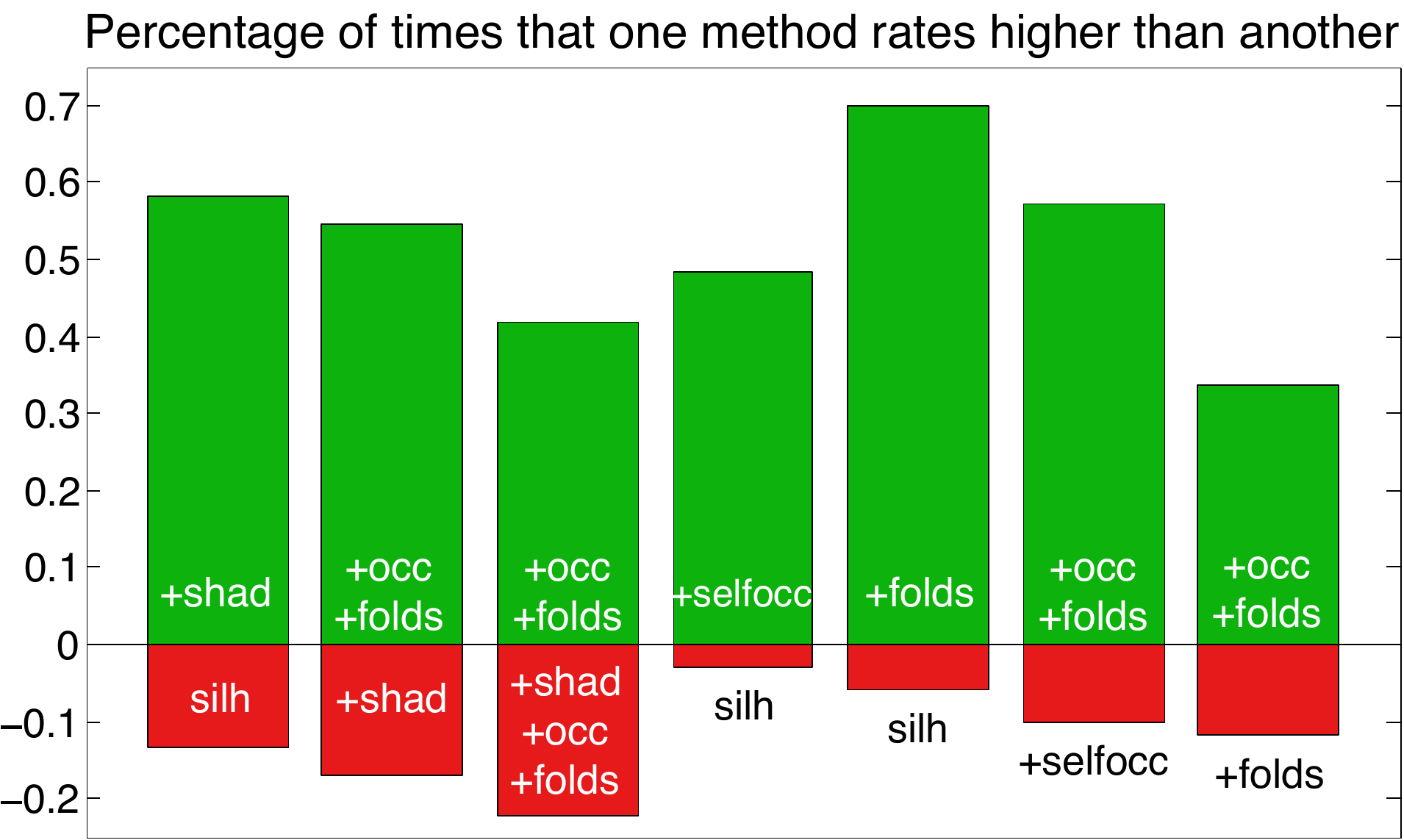}
\caption{The percentage that one algorithm rates higher (green color) or lower (red color) than another. Each column shows the result of one algorithm pair. For example, for the left most column, +shading was rated above silh approximately 60\% of the time, below silh about 10\% of the time, and rated the same otherwise. Shading seems to help when accompanied by with a silhouette cues, but when additional boundary cues are present, shading tends to produce more artifacts than improvements.  We also see a strong improvement from combining fold and occlusion contours.}
\label{fig:fraction}
\end{figure}
%-----------------------------------------------

%%%%%%
\subsection{Shape quality}
\label{sec:eval:quality}
Our experiments examine shape quality perceived by people (through a user study) and computers (ground truth comparison). The goal of these experiments is to find a common set of cues, or shape reconstruction algorithm(s), that consistently report the best shape.

\boldhead{Qualitative rating on VOC}
The qualitative rating portion of the user study collected subjects' ratings for each of the six shape reconstruction algorithms. We designed an interface (Fig.~\ref{fig:interface1}) that displays the visualization of the six shape estimation results side by side on the screen and allows participants to rate the quality of each shape estimation result from scale 1 (very poor) to 5 (very good). The $17$ class $\times$ $10$ instance results are shuffled and divided into 5 groups. Each participant rated an entire group. Additional images from the study are displayed in Fig~\ref{fig:results}.

Our results indicate that +occ+folds is the most appealing reconstruction method to humans, followed closely by +folds and +shading+occ+folds. Figure~\ref{fig:summary} shows the averaged rating score grouped by algorithm; where a higher average rating indicates a better shape. In every case, as intuition suggests, adding more geometric cues leads to a more preferable shape. For an algorithm-by-algorithm comparison, we plot the percentage of times that one algorithm was rated higher than another (Figure~\ref{fig:fraction}). Here, we see geometric cues (other than silh) were consistently preferred over shading cues; in one example, +occ+folds was rated higher than +shading+occ+folds about 40\% of the time.

%\begin{figure}[h]
%\centering
%\includegraphics[width=0.95\linewidth]{fig/userstudy/rating_vis2}
%\caption{Percentage of the ratings stacked for each algorithm.}
%\label{fig:stack}
%\end{figure}

%\begin{figure*}[h]
%\centering
%\includegraphics[width=0.95\linewidth, height=0.2\linewidth]{fig/userstudy/rating_vis4}
%\caption{Stacked rating percentage for each object category (only show three algorithms) \todo{make full page width}.}
%\label{fig:classstack}
%\end{figure*}

%\begin{figure}[h]
%\centering
%\includegraphics[width=0.95\linewidth]{fig/userstudy/rating_vis6}
%\caption{Averaged rating score for each category (only show four algorithms).}
%\label{fig:classsummary}
%\end{figure}

%\begin{figure}[h]
%\centering
%\includegraphics[width=0.95\linewidth]{fig/userstudy/rating_vis5}
%\caption{}
%\label{fig:onevsone}
%\end{figure}

\boldhead{Ground truth comparison}
Using ground truth shapes available from the MIT Intrinsic Image dataset~\cite{grosse09intrinsic}, we analyze our shape reconstructions using established errors metrics. We report results for both a surface normal-based error metric, $N$-MSE~\cite{barron2012eccv}, as well as for a depth-based error metric, $Z$-MAE~\cite{barron2012cvpr}. $N$-MSE is computed as the mean squared error of the difference in normal orientation (measured in radians), and $Z$-MAE is the translation-invariant absolute error of of depth. Both metrics are averaged per-pixel, over the entire dataset of 20 objects. We also ran the same comparison, but substituted Barron and Malik's learned weights on the MIT dataset for our binary weights ($\delta_{{}^*}$); these results are in the $N$-MSE${}^\dagger$ and $Z$-MAE${}^\dagger$ columns:\\

\centerline{
\begin{tabular}{| l | c | c | c | c |} \hline
& \footnotesize $N$-MSE & \footnotesize $Z$-MAE & \footnotesize $N$-MSE${}^\dagger$ &\footnotesize $Z$-MAE${}^\dagger$ \\ \hline
{\footnotesize \bf silh} & 0.573 & 25.533 & 0.521 & 25.637 \\ \hline
{\footnotesize \bf +selfocc} & 0.565 & 25.198 & 0.498 & 25.342 \\ \hline
{\footnotesize \bf +folds} & 0.496 & 25.562 & 0.501 & 25.400 \\ \hline
{\footnotesize \bf +occ+folds} & 0.487 & 25.161 & 0.482 & 24.983 \\ \hline
{\footnotesize \bf +shading} & 0.874 & 38.968 & 0.310 & 25.793 \\ \hline
{\footnotesize \bf +shading+occ+folds} & 0.574 & 27.379 & 0.350 & 24.492  \\ \hline
\end{tabular}
\vspace{4mm}
}

We observe that adding geometric cues generally increase quantitative performance. One notable exception is in the $N$-MSE${}^\dagger$ column, where +shading alone performs the best. This is almost certainly because all +shading reconstruction has been trained on the MIT dataset, whereas several parameters for the other algorithms have not been (e.g. $\delta_{\mathit{selfocc}}, \delta_{\mathit{fold}}$). Surprisingly, using binary weights (as in the $N$-MSE and $Z$-MAE columns) results in significantly worse +shading performance, but the geometric-based algorithms are largely unaffected. However, for non-MIT reconstructed shapes (e.g. VOC), using binary weights versus the learned weights weights gave perceptually similar results, possibly indicating that these metrics are sensitive to different criteria than human perception.

\begin{figure}[t]
\centering
\includegraphics[width=.9\linewidth]{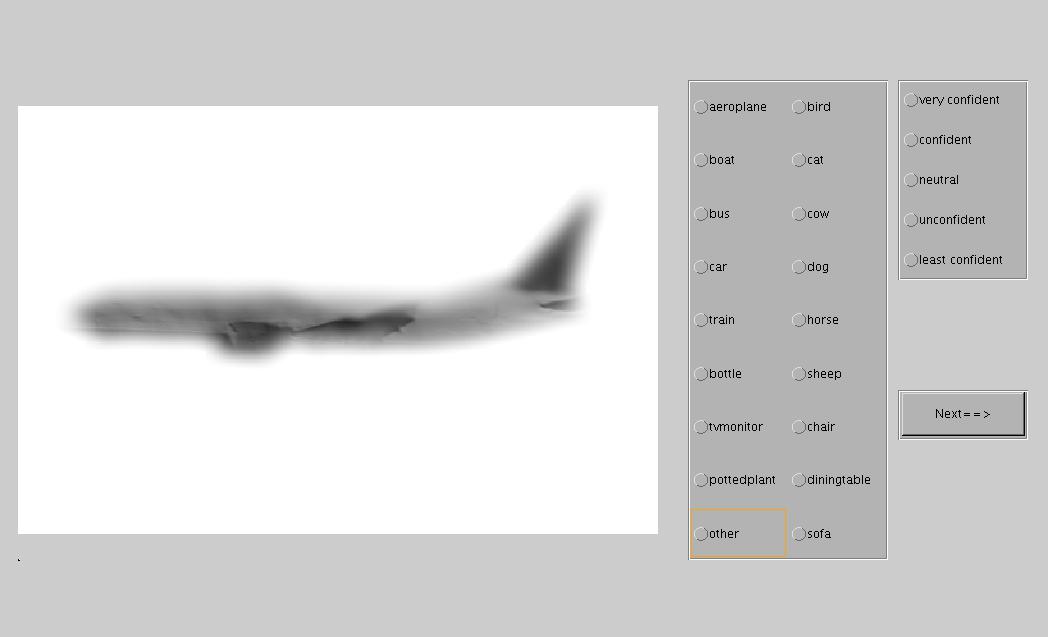}\\
\includegraphics[width=.9\linewidth]{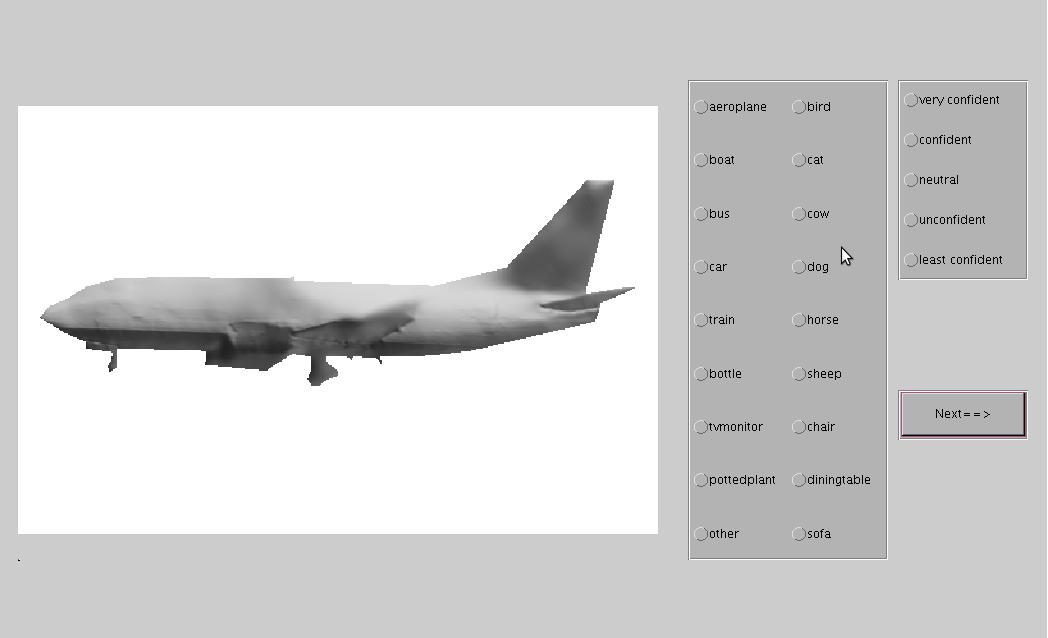}
\caption{User study interface for shape-based recognition. Participants are asked to recognize the object category using shape alone, estimated with one of the six algorithms. For each trial, a ``masked'' view (top) is displayed first to deter silhouette-based recognition, followed by the unmasked view (bottom).
\vspace{-2mm}
}
\label{fig:interface2}
\end{figure}

%%%%%%
\subsection{Shape for object recognition}
\label{sec:eval:recog}

We are also interested in how well our shapes convey the object that has been reconstructed. Here, we describe experiments that gauge this task both through a human recognition study  and computer recognition algorithms.

In the second task, we asked users to identify the object class based on the reconstructed shape alone. Our hypothesis is that higher class-recognition indicates better shape quality. We also consider that object silhouette could be a dominating factor for recognition; to reduce this factor, we show a  silhouette-masked view of each result first (Fig.~\ref{fig:interface2} left); and then show the result without masking (Fig.~\ref{fig:interface2} right). The task is evenly divided into 7 groups. Each group is assigned to one participant, who will go over all of the 170 objects in our test set. For each object, only one of the 7 results (generated by random permutation) is shown to one participant. The participants are also asked to rate their level of confidence on a scale from 1 (least confident) to 5 (most confident).

\begin{figure}[t]
\centering
\includegraphics[width=\linewidth]{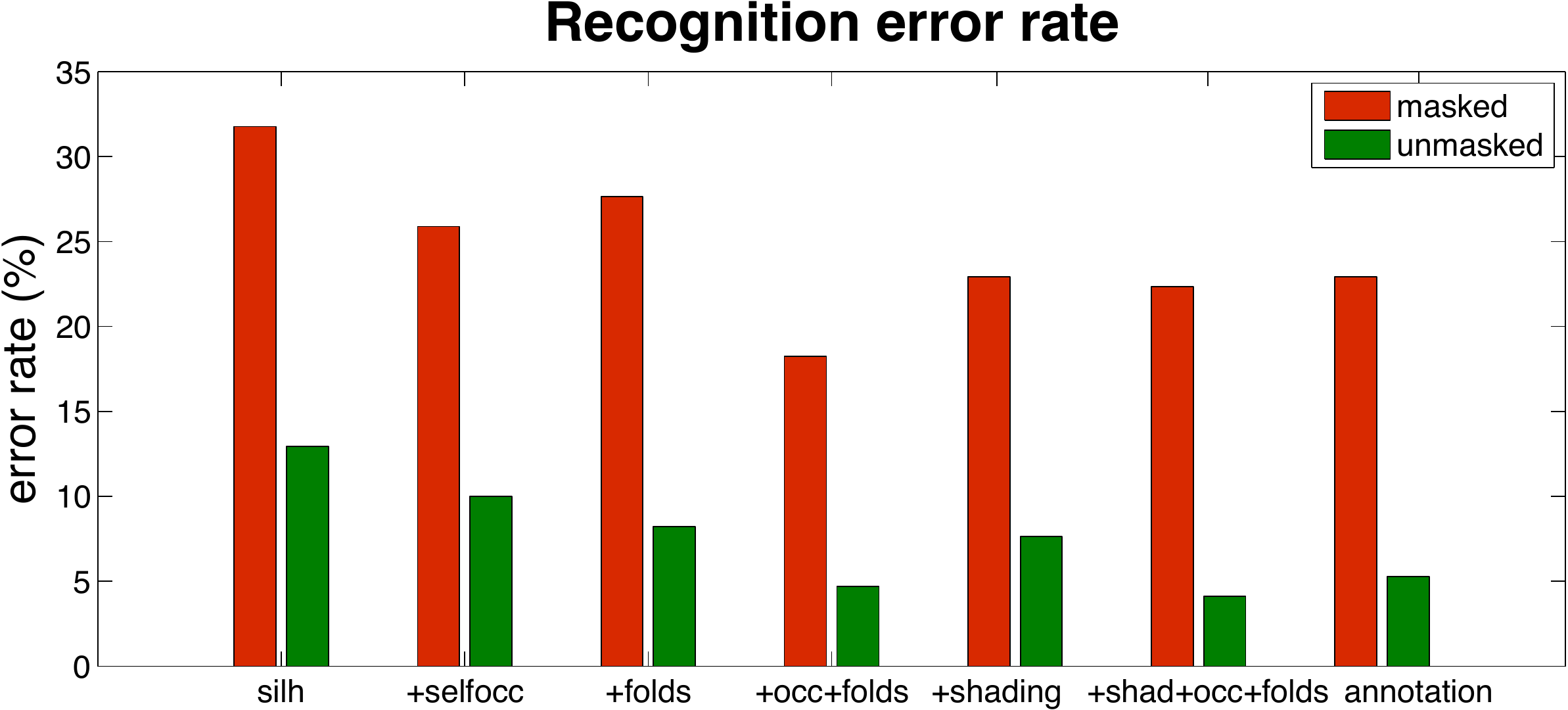}
\caption{Recognition error rate (as judged by participants in our user study) for each algorithm using the silhouette-masked and unmasked (unaltered) images.}
\label{fig:errorrate}
\end{figure}

Figure~\ref{fig:errorrate} displays the recognition error rate for each algorithm. For each algorithm, the left bar shows the result from the masked view; the right bar shows that result from the unmasked view. In the masked view, +occ+folds yields the lowest recognition error, consistent with qualitative rating portion of our user study. In the unmasked view, +shading+occ+folds performs the best, closely followed by +occ+folds.

%\begin{figure}[h]
%\centering
%\includegraphics[width=0.95\linewidth]{fig/userstudy/recog_vis2}
%\caption{Recognition confidence score grouped by the type of results.}
%\label{fig:confidence}
%\end{figure}

%%%%%%
\boldhead{Automatic recognition}
We evaluate the shapes by performing classification on the depth maps.  Outside the image, the depth is set to 0.  Since the heights inside objects are set to be fairly high, this ensures that there is a large edge at the contour.  To provide some invariance to specifics of classification methods or features, we run classification using two methods. We use a PHOW feature from~\cite{bosch2007image} and a the Pegasos SVM solver~\cite{shalev2007pegasos} with homogeneous kernel mapping~\cite{vedaldi2012efficient} as a baseline classifier (all available from VLFeat~\cite{vedaldi10vlfeat}).  It is motivated by a similar method in \cite{silberman11indoor} used to classify objects in Kinect images.  For another method, we use the RGB-D kernel match descriptors of \cite{bo2010kernel,bo2011depth} for which code is available.  Leave one out cross validation is used to determine the accuracy of classification on each reconstruction as well as {rgb}, {rgb+occ+folds}, and {rgb+shading+occ+folds} for the kernel matching method to determine if shape and shading cues add information compared to RGB alone.

Table~\ref{tab:recog} shows classifications results for each of the metrics. Our classification accuracy results are slightly different than the human ratings, although there are some similar trends.  {+occ+folds} still appears to be one of the best, though it is beaten in this case by {+selfocc}. Also expected, {+shading+occ+folds} outperforms {+shading}. The ordering of the remaining reconstructions is less consistent across the two classifiers, therefore it is difficult to draw any strong conclusions. It is interesting that {+folds} performed poorly but was rated highly by our test subjects; this likely implies that the features used do not make use of the information available from folds.

We also show the results of an RGB classifier using \cite{bo2011depth}. While state of the art classification on VOC2012 is roughly 70\%, we see only 55\% due to the constrained dataset (few examples per class).  The shape reconstructions increase the accuracy of the result, but this could be partially due to the mask provided by the height which is not available in the RGB only method.

%The ultimate order is

%bo2011depth
%rgb+occ+folds
%+selfocc
%+occ+folds
%rgb+shading+occ+folds
%rgb
%+shading+occ+folds
%+folds
%+shading
%silh+ss

%PHOW
%+selfocc
%+occ+folds
%+folds
%silh+ss
%+shading+occ+folds
%+shading

%%%%%%%%%%%%%%%%%%%%%%%%%%%%%%%%%%%%%%%%%%%%%%%%
\section{Conclusion}
We demonstrate a simple and extensible technique for reconstructing shape from images, resurrecting highly informative cues from early vision work. Our method itself is an extension of Barron and Malik's~\cite{barron2012eccv} reconstruction framework, and we show how additional cues can be incorporated in this framework to create improved reconstructions.

Through our experiments, we have shown the necessity of considering cues that  go beyond typical shape-from-shading constraints. In almost every task we assessed, using more geometric cues gives better results.  For human-based tasks, shading cues seem to help when applied with to silhouette cues (+shading consistently outperforms silh), but adds little information once additional boundary cues are incorporated (+occ+folds performs similarly to +shading+occ+folds); see Figs~\ref{fig:summary} and~\ref{fig:errorrate}. Interestingly, when the boundary is not available for viewing, +occ+folds performs better than +shading+occ+folds (Fig~\ref{fig:errorrate}; masked errors), and shading cues seem to have an adverse effect on automatic recognition algorithms (Table~\ref{tab:recog}).
%Shading cues help in qualitative tasks involving humans  (in both shape quality, Fig~\ref{fig:summary}, and recognition, Fig~\ref{fig:errorrate}), but only when the silhouette is available. However, shading cues seem to have an adverse effect on automatic recognition algorithms (Table~\ref{tab:recog}).
As far as we know, our experiments are the first to evaluate reconstruction methods on consumer photos (e.g. PASCAL VOC).

\begin{table}
\centering
%\begin{tabular}{|c|c|c|c|c|c|c|c|c|c|}
%\hline
%classifier&rgb & rgb+occ+folds & rgb+shading+occ+folds & +shading & +shading+occ+folds & silh & +selfocc & +folds & +occ+folds \\\hline
%RGB-D kernel~\cite{bo2011depth} & 55.29 & 70.00 & 62.35 & 48.24 & 52.94 & 47.06 & 65.88 & 51.76 & 65.88 \\\hline
%VLFeat~\cite{vedaldi10vlfeat} &-&-&-& 41.76 & 42.94 & 45.29 & 54.12 & 47.65 & 51.76 \\\hline
%\end{tabular}
\begin{tabular}{| l | c | c |} \hline
& \small RGB-D kernel~\cite{bo2011depth} & \small VLFeat~\cite{vedaldi10vlfeat} \\ \hline
\small \bf rgb & 55.29 & - \\ \hline
\small \bf +occ+folds+rgb & \bf 70.00 & - \\ \hline
\small \bf +shading+occ+folds+rgb & 62.35 & - \\ \hline
\small \bf +shading & 48.24 & 41.76 \\ \hline
\small \bf +shading+occ+folds & 52.94 & 42.94 \\ \hline
\small \bf silh & 47.06 & 45.29 \\ \hline
\small \bf +selfocc & 65.88 & \bf 54.12 \\ \hline
\small \bf +folds & 51.76 & 47.65 \\ \hline
\small \bf  +occ+folds & 65.88 & 51.76 \\ \hline
\end{tabular}
\vspace{0mm}
\caption{Average recognition accuracy for different sets of features using existing, automatic recognition methods. rgb implies that image appearance was used as a feature (row 1), and compared against rgb+depth (rows 2 and 3), as well as using depth alone (remaining rows). As one might expect, adding geometric features to the existing rgb information improves recognition accuracy, and shape tends to be more revealing than appearance alone. VLFeat offers only depth classification, hence the missing entries.
\vspace{-2mm}
}
\label{tab:recog}
\end{table}

One interesting observation from our experiments is that our shading cues tend to confound boundary cues; e.g. +occ+folds outperforms +shading+occ+folds in each task except (unmasked) human recognition (Sec ~\ref{sec:eval:recog}). It seems counterintuitive that incorporating shading information would degrade reconstructions, and we offer several possible causes. Foremost is the fact that we weight all terms equally, whereas learning these weights from ground truth will lead to better shading reconstructions (evidenced especially by our quantitative results on the MIT Intrinsic dataset in Sec~\ref{sec:eval:quality}). Second, this observation may be in part due to the inherent assumptions of existing shape-from-shading algorithms, including our own (e.g. 2.5D shape, orthographic camera, Lambertian reflectance, and smooth and infinitely distant illumination). Our tests use real  images from PASCAL, and some contain significant perspective, as well as complex reflectance and illumination. Relaxing these assumptions, as well as developing and enforcing stronger shape priors, are difficult but interesting problems for future research.

Our evaluations show that self occlusion and fold cues are undoubtedly helpful, and most importantly, point in many directions for improving existing shape reconstruction algorithms. Extracting boundary cues, such as folds and self occlusions, automatically from photographs is a logical next step. It is also evident that shape-from-shading algorithms can be improved by incorporating additional geometric cues, and additional research should go into extending shape-from-shading to real world (rather than lab) images. In terms of reconstructing shapes, considering perspective projections (rather than orthographic) may help, as well as extending surface representations beyond 2.5D and into 3D. By exploring these directions, we believe significant steps can be taken in the longstanding vision goal of reconstructing shape in the wild.

%%%%%%%%%%%%%%%%%%%%%%%%%%%%%%%%%%%%%%%%%%%%%%%%
\section*{Acknowledgements} This research was supported in part by the NSF GRFP (KK \& JB), NSF Award IIS 0916014 and ONR MURI Awards N00014-10-10933/N00014-10-10934.

%%%%%%%%%%%%%%%%%%%%%%%%%%%%%%%%%%%%%%%%%%%%%%%%
\renewcommand{\baselinestretch}{.93}
{
\small
\bibliographystyle{ieee}
\bibliography{shapestudy}

\begin{thebibliography}{10}\itemsep=-1pt

\bibitem{barron2012eccv}
J.~T. Barron and J.~Malik.
\newblock Color constancy, intrinsic images, and shape estimation.
\newblock In {\em ECCV}, 2012.

\bibitem{barron2012cvpr}
J.~T. Barron and J.~Malik.
\newblock Shape, albedo, and illumination from a single image of an unknown
  object.
\newblock In {\em CVPR}, 2012.

\bibitem{bo2010kernel}
L.~Bo, X.~Ren, and D.~Fox.
\newblock Kernel descriptors for visual recognition.
\newblock {\em NIPS}, 2010.

\bibitem{bo2011depth}
L.~Bo, X.~Ren, and D.~Fox.
\newblock Depth kernel descriptors for object recognition.
\newblock In {\em Intelligent Robots and Systems (IROS)}, pages 821--826. IEEE,
  2011.

\bibitem{bosch2007image}
A.~Bosch, A.~Zisserman, and X.~Muoz.
\newblock Image classification using random forests and ferns.
\newblock In {\em ICCV}. IEEE, 2007.

\bibitem{felzenszwalb2010pami}
P.~F. Felzenszwalb, R.~B. Girshick, D.~McAllester, and D.~Ramanan.
\newblock Object detection with discriminatively trained part based models.
\newblock {\em IEEE TPAMI}, 32(9):1627--1645, 2010.

\bibitem{ferrari:hal-00204002}
V.~Ferrari, L.~Fevrier, F.~Jurie, and C.~Schmid.
\newblock {Groups of Adjacent Contour Segments for Object Detection}.
\newblock {\em IEEE TPAMI}, 30(1):36--51, Jan. 2008.

\bibitem{grosse09intrinsic}
R.~Grosse, M.~K. Johnson, E.~H. Adelson, and W.~T. Freeman.
\newblock Ground-truth dataset and baseline evaluations for intrinsic image
  algorithms.
\newblock {\em ICCV}, 2009.

\bibitem{hoiem2012eccv}
D.~Hoiem, Y.~Chodpathumwan, and Q.~Dai.
\newblock Diagnosing error in object detectors.
\newblock In {\em ECCV}, 2012.

\bibitem{koenderink1984does}
J.~Koenderink.
\newblock {What does the occluding contour tell us about solid shape}.
\newblock {\em Perception}, 1984.

\bibitem{MalikThesis}
J.~Malik.
\newblock {\em Interpreting line drawings of curved objects}.
\newblock PhD thesis, Stanford University, Stanford, CA, USA, 1986.

\bibitem{MalikMaydan}
J.~Malik and D.~Maydan.
\newblock Recovering three-dimensional shape from a single image of curved
  objects.
\newblock {\em IEEE TPAMI}, 11(6):555--566, June 1989.

\bibitem{Mundy06}
J.~L. Mundy.
\newblock Object recognition in the geometric era: A retrospective.
\newblock In {\em Toward Category Level Object Recognition}, pages 3--29.
  Springer, 2006.

\bibitem{Roberts}
L.~G. Roberts.
\newblock {\em Machine Perception of Three-Dimensional Solids}.
\newblock Outstanding Dissertations in the Computer Sciences. Garland
  Publishing, New York, 1963.

\bibitem{shalev2007pegasos}
S.~Shalev-Shwartz, Y.~Singer, and N.~Srebro.
\newblock Pegasos: Primal estimated sub-gradient solver for svm.
\newblock In {\em ICML}, pages 807--814. ACM, 2007.

\bibitem{silberman11indoor}
N.~Silberman and R.~Fergus.
\newblock Indoor scene segmentation using a structured light sensor.
\newblock In {\em ICCV - Workshop on 3D Representation and Recognition}, 2011.

\bibitem{vedaldi10vlfeat}
A.~Vedaldi and B.~Fulkerson.
\newblock Vlfeat -- an open and portable library of computer vision algorithms.
\newblock In {\em Proceedings of the 18th annual {ACM} International Conference
  on Multimedia}, 2010.

\bibitem{vedaldi09iccv}
A.~Vedaldi, V.~Gulshan, M.~Varma, and A.~Zisserman.
\newblock Multiple kernels for object detection.
\newblock In {\em ICCV}, 2009.

\bibitem{vedaldi2012efficient}
A.~Vedaldi and A.~Zisserman.
\newblock Efficient additive kernels via explicit feature maps.
\newblock {\em IEEE TPAMI}, 34(3):480--492, 2012.

\end{thebibliography}
}
\renewcommand{\baselinestretch}{.95}

\newpage

%%%%%%%%%%%%%%%%%%%%%%%%%%%%%%%%%%%%%%%%%%%%%%%%
\section*{Appendix: Fold constraint implementation}
Consider the $(i)$th point on the contour $C$, parametrized by position $\mathbf{p} = [ \mathbf{p}_x, \, \mathbf{p}_y]$ and tangent vector $\mathbf{u}= [ \mathbf{u}_x,\, \mathbf{u}_y]$, both on the image plane. The sign of the tangent vector is arbitrary. Let us define a vector perpendicular to each tangent vector: $\mathbf{v} = [-\mathbf{u}_y, \, \mathbf{u}_x]$. By default, this fold is convex --- folded in the direction of negative $Z$. To construct a concave fold, we flip the sign of $\mathbf{v}$. With this parametrization, we can find the positions of the points to the left and right of the point in question relative to the contour:
\begin{eqnarray}
\mathbf{p}^{\ell} = [ \round{ \mathbf{p}_x + \mathbf{v}_x}, \, \round{ \mathbf{p}_y + \mathbf{v}_y} ] \\
\mathbf{p}^r = [ \round{ \mathbf{p}_x - \mathbf{v}_x}, \, \round{ \mathbf{p}_y - \mathbf{v}_y} ]
\end{eqnarray}
%\begin{eqnarray}
%\mathbf{p}^{\ell}_x = \round{\mathbf{p}_x + \mathbf{v}_x}, \quad \mathbf{p}^{\ell}_y = \round{ \mathbf{p}_y + \mathbf{v}_y} \\
%\mathbf{p}^{r}_x = \round{ \mathbf{p}_x - \mathbf{v}_x}, \quad \mathbf{p}^{r}_y = \round{ \mathbf{p}_y - \mathbf{v}_y}
%\end{eqnarray}
Given a normal field $N$ we compute the normal of the surface at these ``left'' and ``right'' points:
\begin{eqnarray}
N^{\ell} = [ N_x( \mathbf{p}^{\ell}_x, \mathbf{p}^{\ell}_y), \, N_y( \mathbf{p}^{\ell}_x, \mathbf{p}^{\ell}_y), \, N_z( \mathbf{p}^{\ell}_x, \mathbf{p}^{\ell}_y) ] \\
N^{r} = [N_x( \mathbf{p}^{r}_x, \mathbf{p}^{r}_y), \, N_y(\mathbf{p}^{r}_x, \mathbf{p}^{r}_y), \, N_z( \mathbf{p}^{r}_x, \mathbf{p}^{r}_y) ]
\end{eqnarray}
%\begin{eqnarray}
%\mathbf{n}^{\ell}_x = N_x(p^{\ell}_x, p^{\ell}_y), \quad n^{\ell}_y = N_y(p^{\ell}_x, p^{\ell}_y), \quad n^{\ell}_z = N_z(p^{\ell}_x, p^{\ell}_y) \\
%\mathbf{n}^{r}_x = N_x(p^{r}_x, p^{r}_y), \quad n^{r}_y = N_y(p^{r}_x, p^{r}_y), \quad n^{r}_z = N_z(p^{r}_x, p^{r}_y)
%\end{eqnarray}
Consider $c$, the dot product of $[\mathbf{u}_x, \mathbf{u}_y, 0]$ with the cross-product of $\mathbf{n}^{\ell}$ and $\mathbf{n}^r$:
\begin{eqnarray}
c =  \mathbf{u}_x ( N^{\ell}_y N^r_z - N^{\ell}_z N^r_y) + \mathbf{u}_y ( N^{\ell}_z N^{r}_x - N^{\ell}_x N^r_z)
\end{eqnarray}
If $c=1$, then the cross product of the surface normals on both sides of the contour is exactly equal to the tangent vector, and the surface is therefore convexly folded in the direction of the contour. If $c=-1$, then the surface is folded and concave. Of course, If the sign of the contour, and therefore of the $\mathbf{v}$ vector, is flipped, then $c=1$ when the surface is concavely folded, etc. Intuitively, to force the surface to satisfy the fold constraint imposed by the contour, we should force $c$ to be as close to $1$ as possible. This is the insight used in edge constraint of the shape-from-contour algorithm in \cite{MalikMaydan}. But constraining $c=1$ is not appropriate for our purposes, as it ignores the fact that $\mathbf{u}$ and therefore $\mathbf{v}$ lie in an image plane, while the true tangent vector of the contour may not be parallel to the image plane. To account for such contours, we will therefore penalized $c$ for being significantly smaller than $1$. More concretely, we will minimize the following cost with respect to each contour pixel:
\begin{eqnarray}
f_{\mathit{fold}}(N(Z)) = \sum_{i \in C} \max(0, \epsilon - \supi{c}),
\end{eqnarray}
where $\epsilon = {1 \over \sqrt{2}}$. This is a sort of $\epsilon$-insensitive hinge loss which allows for fold contours to be oriented as much as $45^{\circ}$ out of the image plane. In practice, the value of $\epsilon$ effects how sharp the contours produced by the fold-constraint are --- $\epsilon=0$ is satisfied by a flat fronto-parallel plane, and $\epsilon=1$ is only satisfied by a perfect fold whose crease is parallel with the image plane. In our experience, $\epsilon = {1 \over \sqrt{2}}$ produces folds that are roughly $90^{\circ}$, and which look reasonable upon inspection.

\end{document}